%% file: main.tex
\documentclass[sigconf]{acmart}
\AtBeginDocument{%
  }

\setcopyright{acmlicensed}
\copyrightyear{2018}
\acmYear{2018}
\acmDOI{XXXXXXX.XXXXXXX}
\acmConference[Conference acronym 'XX]{Make sure to enter the correct
  conference title from your rights confirmation email}{June 03--05,
  2018}{Woodstock, NY}
\acmISBN{978-1-4503-XXXX-X/2018/06}

\usepackage{enumitem}
\usepackage{pifont}
\usepackage{multirow}
\usepackage{makecell}
\usepackage{adjustbox}
\usepackage{tikz}
\usepackage{pgfplots}
\usepackage{pgfplotstable}
\usepackage{float}
\usepackage{booktabs}
\usepackage{graphicx}
\usepackage{subcaption}
\usepackage{pdfpages}
\usetikzlibrary{patterns}
\pgfplotsset{compat=1.3}
\newcommand{\cmark}{\ding{51}}
\newcommand{\xmark}{\ding{55}}
\usepackage{dblfloatfix}
\usepackage{xcolor}
\usepackage[dvipsnames]{xcolor}

\copyrightyear{2026}
\acmYear{2026}
\setcopyright{cc}
\setcctype{by}
\acmConference[KDD '26]{Proceedings of the 32nd ACM SIGKDD Conference on Knowledge Discovery and Data Mining V.2}{August 09--13, 2026}{Jeju Island, Republic of Korea}
\acmBooktitle{Proceedings of the 32nd ACM SIGKDD Conference on Knowledge Discovery and Data Mining V.2 (KDD '26), August 09--13, 2026, Jeju Island, Republic of Korea}
\acmDOI{10.1145/3770855.3818077}
\acmISBN{979-8-4007-2259-2/2026/08}
\begin{document}



\title{MLaGA: Multimodal Large Language and Graph Assistant}

\author{Dongzhe Fan}
\affiliation{%
  \institution{New York University Shanghai}
  \city{Shanghai}
  \country{China}
}
\email{df2362@nyu.edu}

\author{Jiajin Liu}
\affiliation{
 \institution{New York University}
 \city{Brooklyn}
 \state{NY}
 \country{USA}
}
\email{jiajinliu@nyu.edu}

\author{Yi Fang}
\affiliation{
  \institution{Virginia Polytechnic Institute and State University}
  \city{Blacksburg}
  \state{VA}
  \country{USA}
}
\email{yif@vt.edu}

\author{Djellel Difallah}
\affiliation{
 \institution{New York University Abu Dhabi}
 \city{Abu Dhabi}
 \country{United Arab Emirates}
}
\email{djellel@nyu.edu}

\author{Qiaoyu Tan}
\authornote{Corresponding author}
\affiliation{
 \institution{New York University Shanghai}
 \city{Shanghai}
 \country{China}
}
\email{qiaoyu.tan@nyu.edu}


\begin{abstract}
\input{sections/Abstract}
\end{abstract}




\maketitle
\newcommand\kddavailabilityurl{https://doi.org/10.5281/zenodo.20493431}
\ifdefempty{\kddavailabilityurl}{}{
\begingroup\small\noindent\raggedright\textbf{Resource Availability:}\\
The source code of this paper has been made publicly available at \url{\kddavailabilityurl}.
\endgroup
}

\section{Introduction}
\input{sections/Introduction}

\section{Related Work}
\input{sections/Related_Work}

\section{Problem Formulation}
\input{sections/Problem_Formulation}

\section{Methodology}
\input{sections/Methodology}

\section{Experiments}
\input{sections/Experiments}

\section{Conclusion and Future Work}
\input{sections/Conclusion}


\bibliographystyle{ACM-Reference-Format}
\bibliography{main}

\appendix
\input{sections/Appendix}
\end{document}

%% file: sections/Abstract.tex
Large language models (LLMs) have shown strong potential in graph learning by enabling powerful reasoning and broad generalization. However, existing Graph LLMs remain confined to textual graphs, where node features are represented solely by textual descriptions. This narrow focus overlooks a growing class of multimodal graphs (MMGs), where nodes are associated with multimodal attributes, such as text and images, commonly seen in real-world applications like e-commerce, social networks, and digital artworks. Effectively modeling MMGs presents two key challenges: (\textit{i}) capturing fine-grained cross-modal interactions—e.g., between visual patches and word tokens—while preserving structural dependencies, and (\textit{ii}) achieving unified generalization across diverse tasks and domains within a single model. To address these challenges, we propose \textbf{MLaGA}, a novel \textbf{M}ultimodal Large \textbf{L}anguage \textbf{a}nd \textbf{G}raph \textbf{A}ssistant that serves as the LLM-based foundation model for multimodal graphs. MLaGA comprises two key innovations: (1) the \textit{Structure-Aware Multimodal Aligner (SMA)}, which performs token-level fusion of visual and textual representations via query-driven cross-attention while explicitly maintaining graph topology, delivering high-quality multimodal node representations; and (2) the \textit{Multi-Task Multimodal Graph Instruction Tuning (MMGIT)} framework, which combines structured prompts with a novel cross-task attention mechanism over task-specific projectors to enable scalable instruction tuning across multiple graph tasks, significantly improving cross-task generalization. 
Together, these components endow MLaGA with fine-grained multimodal reasoning and strong generalization across domains and objectives. 
Extensive experiments on 12 diverse MMG benchmarks demonstrate that MLaGA consistently surpasses state-of-the-art baselines on standard node classification and link prediction tasks and multimodal generative tasks, establishing a new foundation for multimodal graph learning while remaining easily extensible to new tasks.

%% file: sections/Introduction.tex
Graphs are ubiquitous and powerful data structures for modeling complex relationships and interactions among entities in a wide range of domains, including social networks~\cite{wu2020comprehensive}, molecular graphs~\cite{zhou2020graph,liu2024moleculargptopenlargelanguage}, and recommendation systems~\cite{wu2022graph}. Traditionally, graph learning has relied on specialized and task-specific models tailored to individual datasets or applications. However, with the emergence of foundation models, the research paradigm has shifted toward the development of Graph Foundation Models (GFMs)-general-purpose models that aim to support a wide range of tasks and domains. Inspired by the success of Large Language Models (LLMs) to reason over graph-structured data, recent works~\cite{chen2024llagalargelanguagegraph,www24graphprompter,zhang2024graphtranslator,tang2024graphgptgraphinstructiontuning,fang2024uniglmtrainingunifiedlanguage} leverage LLMs’ strong contextual reasoning capabilities to perform graph-based tasks, especially by incorporating structural information from graphs into the language modeling process. Nevertheless, these models have largely focused on Text-Attributed Graphs (TAGs), where node features are represented exclusively through textual descriptions. 

In contrast, many real-world graphs are multimodal in nature, where node attributes span multiple modalities such as text and images, resulting in Multimodal Graphs (MMGs)~\cite{epstein2023art,huang2022draw,jin2024instructg2isynthesizingimagesmultimodal,peng2024learningmultimodalgraphssurvey, wei2024gitagraphvisualtextual}. For example, in e-commerce applications, each product node is typically associated with both image and textual content~\cite{zhu2024multimodal}. These heterogeneous signals provide complementary semantic information, making multimodal representation learning essential for fine-grained user profiling and item understanding. However, despite their growing prevalence, existing GFMs are not well-equipped to process MMGs, remaining limited to unimodal textual settings and ignoring the complementary potentials hidden in multimodal data. 

In the literature, some progress has been made in multimodal graph learning~\cite{hu2021mmgcnmultimodalfusiondeep,tao2020mgat,liu_2023_multimodal_graph}. Most existing methods adopt general-purpose pre-trained encoders such as CLIP~\cite{radford2021learningtransferablevisualmodels} to extract visual and textual features for each node in a preprocessing step. While this pipeline is simple and modular, it introduces two major limitations. First, the quality and effectiveness of the learned node representations are tightly coupled with the pre-trained encoders, which may not generalize well to graph-based domains due to domain shift. Second, these methods are typically built upon conventional Graph Neural Networks (GNNs)~\cite{he2025unigraph2learningunifiedembedding, hu2021mmgcnmultimodalfusiondeep}, which, despite their success in modeling local topological structures, lack the expressiveness and generalization capacity needed for large-scale, cross-domain reasoning. These limitations prevent existing approaches from fulfilling the broader goals of foundation models in multimodal graph settings. 

Recent advances in Graph LLMs~\cite{tang2024graphgptgraphinstructiontuning,chen2024llagalargelanguagegraph} demonstrate the potential of combining LLMs with graph structures to improve generalization and task transfer. However, current Graph LLMs are still confined to unimodal text-based graphs and have not yet been extended to support multimodal reasoning. This reveals a significant gap between the capabilities of current multimodal graph learning models and the emerging vision of graph foundation models. To close this gap, there is an urgent need to develop LLM-based multimodal graph foundation models that can integrate heterogeneous node attributes and support robust generalization across diverse domains and tasks in a unified framework.

However, developing such a model introduces several non-trivial challenges.~\ding{192} \textbf{Fine-Grained Structure-Aware Multimodal Alignment.} Existing Graph LLM approaches typically convert visual and textual node attributes into a single feature vector through coarse fusion strategies such as concatenation or global pooling, often leveraging pre-trained multimodal encoders like CLIP~\cite{he2025unigraph2learningunifiedembedding} or fine-tuning them on specific multimodal graphs~\cite{liu_2023_multimodal_graph}. However, relying predominantly on the ``CLS'' token to summarize each modality may ignore essential fine-grained semantic interactions, such as visual patch and work token alignments, that are critical for capturing nuanced semantic relationships. Achieving fine-grained, structure-aware alignment that jointly models intra-node multimodal semantics and inter-node structural dependencies is therefore essential for accurate node representation. 
\ding{193} \textbf{Cross-Task Generalization.} While recent Graph LLMs demonstrate encouraging cross-domain generalization through joint training on multiple datasets, their performance in cross-task scenarios—where the model must handle different types of reasoning objectives such as classification and link prediction—remains under-explored and limited. 
For example, LLaGA~\cite{chen2024llagalargelanguagegraph} achieves notable gains in multi-domain single-task settings but experiences significant degradation when trained across multiple tasks. This limitation stems from the lack of explicit modeling of task-specific reasoning patterns; most existing approaches rely on a shared semantic space that does not distinguish between task objectives. 
Building a single model that can generalize across both domains and tasks in a unified manner remains a core and open challenge in the development of graph foundation models.



To tackle the aforementioned challenges, we present \textbf{MLaGA}—a unified foundation model that extends LLM reasoning to multimodal graphs. MLaGA is built on two complementary components. First, the \textit{Structure-Aware Multimodal Aligner (SMA)} leverages query-driven cross-attention to perform token-level fusion of textual and visual node attributes, enabling semantically rich and structurally consistent node representations. Second, we propose the \textit{Multi-Task Multimodal Graph Instruction Tuning (MMGIT)} framework, which unifies instruction tuning for multiple graph tasks by integrating structured prompts with a cross-task attention mechanism that collaboratively distills knowledge across task-specific projectors. This two-stage framework equips MLaGA with strong generalization capabilities across multiple tasks and domains, achieving state-of-the-art results on a diverse suite of multimodal graph benchmarks.

Our key \textbf{contributions} are summarized as follows:
\begin{itemize}[itemsep=0pt,topsep=0pt,leftmargin=*,label=$\star$]
    \item We introduce \textbf{MLaGA}, the first LLM-based foundation model designed for multimodal graphs, capable of reasoning over heterogeneous visual and textual attributes while generalizing across multiple tasks in a unified architecture. 
    \item We develop the \textbf{Structure-Aware Multimodal Aligner (SMA)}, a fine-grained token-level alignment module that combines learnable query vectors with cross-attention, enabling effective integration of multimodal node attributes and structural context, significantly improving the quality of multimodal node representations.
    \item We propose the \textbf{Multi-Task Multimodal Graph Instruction Tuning (MMGIT)} framework, which supports scalable instruction tuning for multiple graph tasks. MMGIT incorporates structured multimodal prompts and a novel cross-task projector module to promote task-specific reasoning while enabling knowledge sharing across tasks, significantly mitigating performance degradation observed in traditional single-projector fine-tuning. 
    \item We conduct extensive evaluations on twelve multimodal graph datasets from diverse domains—including social networks, e-commerce, book-comment and digital artwork. MLaGA consistently outperforms competitive baselines in both supervised and transfer learning settings, setting a new standard for multimodal graph reasoning with LLMs.
\end{itemize}

%% file: sections/Related_Work.tex
Our work is closely related to the following two research directions.


\textbf{Graph LLMs}. 
Large Language Models (LLMs) have catalyzed significant advancements in the field of graph learning, enabling enhanced capabilities in processing and reasoning over complex, structured data. In terms of the exceptional generative capabilities of LLMs, models like GaugLLM\cite{fang2024gaugl}, advance graph self-supervised learning through augmented node feature by a MoE module. Other than the graph node feature augmentation paradigm, graph LLMs also explore effective frameworks to encode graph structural information into LLMs' semantic space. GraphPrompter\cite{www24graphprompter} encodes graph structures into embeddings, serving as soft prompts to guide LLMs in graph-related tasks. To learn a foundation model for text-attributed graphs, GraphGPT\cite{tang2024graphgptgraphinstructiontuning} incorporates graph structural knowledge into LLMs embedding space through a dual-stage instruction tuning process and Unigraph\cite{he2024unigraph} integrates LLMs with GNNs, leveraging masked graph modeling and instruction tuning to enhance cross-domain generalization. LLaGA\cite{chen2024llagalargelanguagegraph} converts graph nodes into structure-aware sequences and maps them into the LLM embedding space through a trainable projector. Nevertheless, these approaches are confined to text-attributed graphs, neglecting the challenges of graph reasoning in MMGs.


\textbf{Multimodal Graph Models}.  
With the remarkable advancements in Computer Vision, research efforts on graphs are increasingly focusing on MMGs to enhance understanding and reasoning across multiple modalities, extending beyond text alone. MMGCN~\cite{hu2021mmgcnmultimodalfusiondeep} adopts a spectral‐domain, multi‐layer GCN to encode both contextual and multimodal signals in dialogue, while MGAT~\cite{tao2020mgat} augments this design with an attention mechanism that differentially weights neighboring nodes. However, both architectures have been evaluated solely within recommendation settings. Unigraph2~\cite{he2025unigraph2learningunifiedembedding} overcomes this narrow focus by employing a Mixture‐of‐Experts (MoE) network to integrate cross‐modal and cross‐graph node representations, thus extending multimodal graph techniques to conventional graph‐based tasks. MM-GRAPH~\cite{zhu2025mosaicmodalitiescomprehensivebenchmark}  and MAGB~\cite{yan2025graphmeetsmultimodalbenchmarking} conduct extensive experiments to comprehensively evaluate multimodal graph learning across diverse real-world tasks. However, these methods require pre-trained multimodal encoders, which may ignore fine-grained multimodal interactions and cannot generalize across diverse graph learning tasks. Thus, we aim to align different modalities at a finer granularity and empower LLMs to achieve cross-task generalization in multimodal graph reasoning.

\begin{figure*}[h]
\centering
\includegraphics[width=18cm, height=5cm]{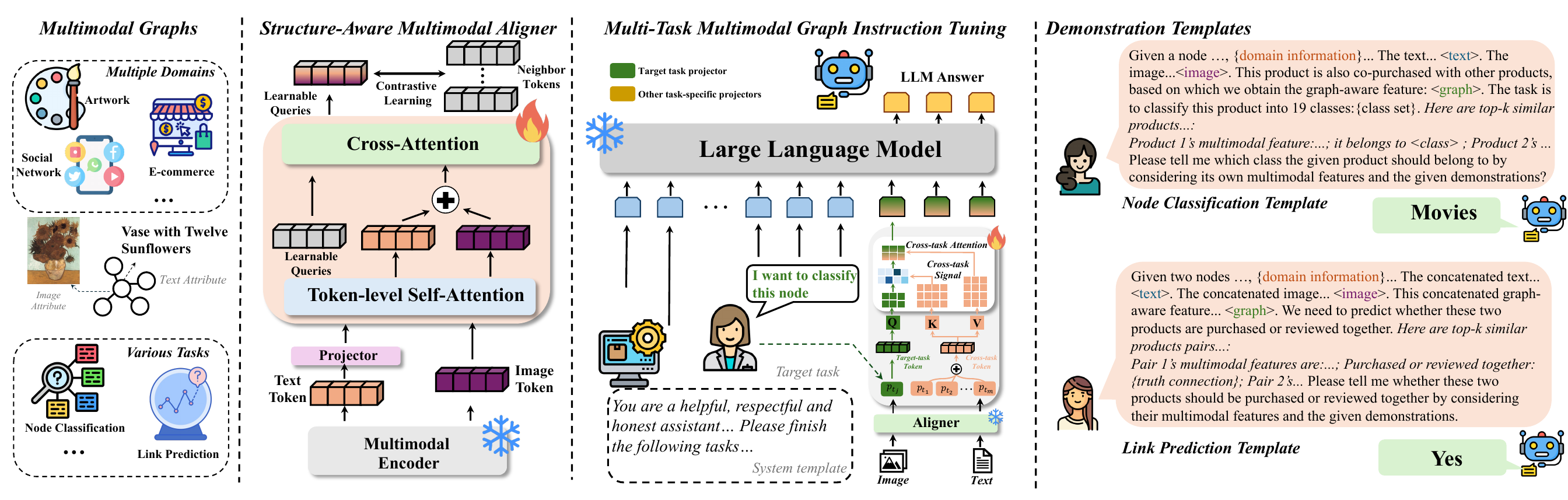}
\caption{\textbf{The overview of MLaGA framework. MLaGA involves a two-stage pre-training framework. (1) MLaGA trains a unified structure-aware multimodal aligner across multiple MMGs using a contrastive graph pre-training objective. (2) MLaGA adopts a Multi-task Multimodal Graph Instruction framework, composed of multiple task‑specific projectors alongside a cross‑task attention module. This design enables the projector to learn both task‑specific expertise and shared cross‑task knowledge, thereby achieving effective cross‑task and cross-domain generalization.}}
\label{fig:mlaga}
\end{figure*}

%% file: sections/Problem_Formulation.tex
\begin{definition}{(Multimodal Graphs (MMGs))}
A multimodal graph is defined as $\mathcal{G}=(\mathcal{V}, \mathcal{E}, \mathcal{I}, \mathcal{T}, \mathcal{Y})$, where $\mathcal{V}$ is the set of nodes, $\mathcal{E}$ the set of edges, $\mathcal{I}$ the collection of images, $\mathcal{T}$ the set of documents, and $\mathcal{Y}$ the label space. Each node $v_i\in \mathcal{V}$ is associated with an image attribute $\mathcal{I}_{v_i}\in\mathcal{I}$ and a textual attribute $\mathcal{T}_{v_i}\in \mathcal{T}$, and may have a label $\mathcal{Y}_{v_i}\in\mathcal{Y}$ for the classification task. If node $v_i$ is connected to node $v_j$, then $(v_i, v_j)\in\mathcal{E}$. 
\end{definition}
An example of an MMG is the Amazon co-purchase network ($\mathcal{G}$), where each node ($v_i\in\mathcal{V}$) represents a product sold on Amazon. Each node is associated with a product image ($\mathcal{I}_{v_i} \in \mathcal{I}$) and a short textual description (e.g., product details or user reviews) ($\mathcal{T}_{v_i} \in \mathcal{T}$) related to this product. 

\textbf{Problem Definition}. Given a multimodal graph $\mathcal{G}$, the objective is to learn a predictive function $f$ that leverages the node-level multimodal attributes ($\mathcal{I},\mathcal{T}$) and the underlying graph structure in $\mathcal{E}$. The trained model should accurately perform tasks such as node classification—predicting a node’s label—and link prediction—estimating connectivity between any two nodes, while also demonstrating strong generalization to previously unseen graphs without requiring additional fine-tuning.

%% file: sections/Methodology.tex

In this section, we introduce the MLaGA framework—a novel approach that extends general-purpose LLMs for reasoning over multimodal graphs (MMGs). In Section~\ref{sec-aligner}, we describe the structure-aware multimodal aligner, which addresses the heterogeneity of node attributes by aligning textual and visual information in the context of graph connectivity. In Section~\ref{sec-mgic}, we present the multi-task multimodal graph instruction tuning framework, which collaboratively extracts complementary information across diverse tasks and domains to advance MLaGA's generalization capabilities for multimodal graph reasoning.

\subsection{Structure-Aware Multimodal Aligner}
\label{sec-aligner}
To address the challenge of multimodal node attributes in MMGs, we propose the \textbf{Structure-Aware Multimodal Aligner (SMA)}. SMA integrates visual and textual node attributes through fine-grained cross-modal attention and aligns the resulting representations with the graph structure. This design enables the model to produce semantically rich and structurally informed node embeddings that are well-suited for downstream LLM-based reasoning. 
As shown in Figure~\ref{fig:mlaga}, SMA comprises three main components: (1) shared token-level self-attention, (2) query-based cross-modal fusion, and (3) structure-aware contrastive pretraining.
\subsubsection{Shared Token-level Self-Attention}
Given an anchor node $v_i$ with image attribute $\mathcal{I}_{v_i}$ and text attribute $\mathcal{T}_{v_i}$, we first employ a pre-trained multimodal encoder (e.g., CLIP~\cite{radford2021learningtransferablevisualmodels}) to obtain initial token-level representations. Assume $\mathbf{H}_{\text{img},v_i} = \phi(I_{v_i}) \in \mathbb{R}^{n_v\times d_i}$ and $\mathbf{H}_{\text{txt},v_i} = g(T_{v_i})\in \mathbb{R}^{n_t\times d_t}$ represent the resulting image patch sequence and text token sequence generated by the visual encoder $\phi(\cdot)$ and textual encoder $g(\cdot)$, respectively. Here, $n_v$ and $n_t$ denote the number of image patches and text tokens, and $d_i$ and $d_t$ are their respective embedding dimensions. After this, we adopt a projection layer to map the text tokens into the same latent space of image tokens for subsequent mutlitmodal fusion: $\mathbf{H}_{\text{text}, v_i} = \text{Proj}(\mathbf{H}_{\text{text}, v_i}) \in \mathbb{R}^{n_t\times d_i}$.
To capture intra-modality contextual dependencies, we apply a shared self-attention module to both image and text sequences:
\begin{equation}
\begin{split}
    \mathbf{H}_{\text{txt},v_i} = \text{ShareAttn}(q, k, v=\mathbf{H}_{\text{txt},v_i})\in \mathbb{R}^{n_t\times d}\\
    \mathbf{H}_{\text{img},v_i} = \text{ShareAttn}(q,k,v=\mathbf{H}_{\text{img},v_i} )\in \mathbb{R}^{n_v\times d},
\end{split}
\end{equation}
$\text{ShareAttn}(\cdot)$ is a standard self-attention layer with multiple heads, which allows each modality to exploit its internal context to emphasize the most informative tokens. This shared attention layer allows both modalities to highlight informative tokens while projecting them into a common latent space of dimension $d$.

\subsubsection{Cross-Modal Fusion with  Q-Queries}
To enable cross-modal interaction and token-level fusion, we introduce a set of learnable query vectors $\mathbf{Q}\in\mathbb{R}^{n_q\times d}$. 
Inspired by~\cite{li2023blip2bootstrappinglanguageimagepretraining}, we randomly initialize $n_q$ learnable queries in each cross-modal layer; these queries interact with the concatenated visual and textual tokens via a cross-attention mechanism:
\begin{equation}
    \mathbf{Q}_{v_i} = \text{CrossAttn}(q=\mathbf{Q}; k,v=\mathbf{H}_{\text{img},v_i}||\mathbf{H}_{\text{txt},v_i}),
    \label{eq-crossattn}
\end{equation}
where $||$ denotes the concatenation operation. $\text{CrossAttn}(\cdot)$ is another attention layer that concatenates the hidden sequences of image and text as key and vector arguments. This results in $\mathbf{Q}_{v_i}\in\mathbb{R}^{n_q\times d}$, a fixed-size multimodal representation for node $v_i$. This design offers two key benefits: (1) Efficient token compression: Eq.~\ref{eq-crossattn} distills fine-grained multimodal signals into a compact sequence suitable for LLM consumption; (2) Node-level adaptability: Each node learns to attend to modality-specific patterns depending on its unique input, improving representation diversity and robustness, which is crucial for graph learning tasks. 


\subsubsection{Structure-Aware Contrastive Pre-training}
While $\mathbf{Q}_{v_i}$ captures multimodal semantics, we further align these embeddings with graph topology through a contrastive pretraining objective. Instead of reconstructing edges via auto-encoder~\cite{kipf2016variational} (which can lead to overfitting~\cite{velivckovic2018deep}), we follow a neighborhood contrastive learning strategy that promotes consistency among structurally related nodes: 
\begin{equation}
\begin{split}
\mathcal{L}_{v_i} = -\sum_{v_u\in \mathcal{N}(v_i)} \log \frac{\exp \left( {\text{sim}(\text{pl}(\mathbf{Q}_{v_i}), \text{pl}(\mathbf{Q}_{v_u}}))/{\tau} \right)}{\sum\limits_{k \in \mathcal{B}} \exp \left( {\text{sim}(\text{pl}(\mathbf{Q}_{v_i}), \text{pl}(\mathbf{Q}_{v_k})})/{\tau} \right)},
\end{split}
\label{conloss}
\end{equation}
where $\mathcal{N}(v_i)$ denotes a set of highly similar neighbors, randomly sampled from $v_i$'s local neighborhood. $\text{sim}(\cdot)$ is cosine similarity, $\tau$ is the temperature parameter, $\text{pl}(\cdot)$ is a pooling function (e.g., mean) over the query tokens. and $\mathcal{B}$ denotes samples in the current mini batch. 
By optimizing Eq.\ref{conloss}, it encourages nodes with similar local structures to generate similar multimodal embeddings. 

\noindent\textbf{Remark}. SMA mitigates modality misalignment by combining token-level aggregation with structure-aware supervision. For example, when the image modality is noisy or is not fully semantically aligned with the accompanying text, the model adaptively prioritizes the more reliable visual or textual tokens. In addition, SMA is both scalable and generalizable: it operates on local subgraphs without relying on full-graph message passing, making it well-suited for universal multimodal pretraining across diverse graph domains.


\subsection{{Multi-Task Multimodal Graph Instruction Tuning}}
\label{sec-mgic}
While instruction tuning has achieved impressive results in both graph~\cite{ye2024languagegraphneeds, sun2025graphiclunlockinggraphlearning, zhao2023graphtextgraphreasoningtext} and visual-language domain~\cite{li2024llavanextinterleavetacklingmultiimagevideo, wang2024qwen2vlenhancingvisionlanguagemodels}, its extension to multimodal graph reasoning remains largely underexplored. A key obstacle lies in enabling strong generalization across diverse graph tasks-such as node classification and link prediction-within a unified instruction-tuned framework. As shown in Table~\ref{tab:main_results}, existing graph LLM efforts are effective in cross-domain transfer when trained on multiple graphs, but often fail to maintain consistent performance across tasks when jointly trained, lagging behind task-specific counterparts.  

To bridge the gap, we propose the \textbf{Multi-Task Multimodal Graph Instruction Tuning (MMGIT)} framework, a principled instruction solution on multimodal graphs that encourages collaborative knowledge sharing across heterogeneous tasks. MMGIT comprises two core components: (1) a multi-task cross-attention projector, which adaptively captures both task-specific signals and transferable patterns across tasks, and (2) a task-wise demonstration selection strategy,  which constructs contextually relevant and structurally grounded prompts to enhance multimodal graph reasoning in LLMs, as elaborated below.
\subsubsection{Multi-task Cross-attention Projector}
The goal of this module is to map multimodal node representations obtained from SMA into LLM-compatible semantic space while capturing both task-specific and shared inductive signals. In traditional graph LLMs~\cite{chen2024llagalargelanguagegraph}, a single projector is typically used to adapt embeddings across tasks. However, this shared projector often leads to suboptimal performance due to conflicting task-specific objectives. We instead propose a multi-task projector architecture that balances specialization and generalization by combining \textit{task-specific projectors} with a \textit{cross-task attention mechanism}.


\noindent\textbf{Task-specific Projectors.} Given a task set $T =\{t_1, t_2, ..., t_m\}$, we allocate a dedicated projector $p_{t_j}$ (i.e., a linear transformation layer) for each task $t_j$. For a node $v_i$ in the context of task $t_j$, the corresponding task-specific representation is computed as:
\begin{equation}
    \mathbf{e}_{v_i}^{t_j} = f_{p_{t_j}}(\mathbf{Q}_{v_i})
\end{equation}
Where $\mathbf{Q}_{v_i}$ is the multimodal representation output by SMA. This design ensures that each task learns representations tailored to its specific objective, mitigating negative transfer and improving task fidelity.

\noindent\textbf{Cross-task Attention Layer.} To capture transferable knowledge across tasks, we incorporate a cross-task attention layer. For each task $t_j$, its task-specific output $\mathbf{e}_{v_i}^{t_j}$ acts as the query, while the outputs from all other projectors are used as keys and values:
\begin{equation}
    \mathbf{e}_{v_i}^{\prime} = \mathrm{CrossAttn}(q=\mathbf{e}_{v_i}^{t_j} ; k,v= \bigl[\mathbf{e}_{v_i}^{t_{j^{\prime}}}\bigr]_{t_{j^{\prime}}\in[T]\setminus\{t_j\}})
\end{equation}
This mechanism allows the model to adaptively borrow shared semantics from auxiliary tasks that may improve performance on the target task. The final enriched task representation is obtained via residual connection and feed-forward refinement:
\begin{equation}
    \hat{\mathbf{e}}_{v_i} = \mathrm{FFN(\mathbf{e}_{v_i}^{\prime})} + \mathbf{e}_{v_i}^{t_j}
\end{equation} 
By leveraging both task-specific and cross-task representations, the proposed framework can better balance specialization and generalization across various tasks, leading to more robust performance, as shown in Table~\ref{tab:ablation_ATM}.

\subsubsection{Task-wise Demonstration Selection}
As illustrated in Figure ~\ref{fig:mlaga}, to effectively fine-tune LLMs for multimodal graph learning, we construct input prompts that incorporate graph-structured demonstrations. Given the input length constraints of LLMs and the potential noise present in raw graph structures, it is essential to carefully select concise and informative demonstration sets tailored to the specific requirements of each task. In this work, we mainly focus on two fundamental graph learning tasks: node classification and link prediction, following prior graph LLM studies~\cite{tang2024graphgptgraphinstructiontuning, he2025unigraph2learningunifiedembedding, chen2024llagalargelanguagegraph}. These tasks are representative of the broader landscape of graph reasoning problems. In what follows, we describe how to design task-specific prompt templates for each of them.

\noindent\textbf{Node Classification Demonstration}. 
For node classification task, a natural strategy is to construct the demonstration set $\mathcal{D}(v_i)$ by selecting nodes that are semantically and structurally similar to the anchor node $v_i$. However, relying solely on multimodal embedding similarity overlooks important structural context. To address this, we integrate both feature-based similarity and graph topology using a hybrid scoring mechanism. Specifically, we compute the Personalized PageRank (PPR) scores $\mathbf{S}_{i, j}$ between node $v_i$ and each candidate node $v_j$ to capture their structural proximity as follows:
\begin{equation}
    \label{eq:7}
    S = \alpha \mathbf{p} + (1 - \alpha) \mathbf{A}^\top \mathbf{r},
\end{equation}
where $\alpha$ is the teleport probability controlling the trade-off between personalization and structural diffusion, $\mathbf{A}$ is the normalized adjacency matrix, $\mathbf{p}$ is the personalization vector, and $r$ is the PageRank score vector. To capture both semantic and structural relevance, we adopt a hybrid selection approach: we first rank candidate nodes based on cosine similarity to construct a semantic shortlist, and then select the top-$k$ nodes with the highest PPR scores from this list to form the final demonstration set $\mathcal{D}(v_i)$, ensuring that each demonstration encodes both semantic and topological relevance.

\noindent\textbf{Link Prediction Demonstration}.
Unlike node classification, selecting effective demonstrations for link prediction is more challenging, as it involves reasoning over a pair of nodes rather than a single one. To address this, we adopt a neighborhood-based edge selection strategy. Given an anchor edge $(v_i, v_j)$, we identify structurally relevant demonstration edges by exploring the intersecting or shared neighborhoods of $v_i$ and $v_j$. Specifically, we define the demonstration set as: $\mathcal{E}_{\text{demo}}^{v_i,v_j} = \left\{(u, v) \;\middle|\; u \in \mathcal{N}(v_i) \cap \mathcal{N}(v_j) \;\text{or}\; v \in \mathcal{N}(v_i) \cap \mathcal{N}(v_j) \right\}$. This approach ensures that the selected edges are contextually relevant to both endpoints of the anchor link, effectively capturing local structural patterns that are indicative of link formation. To avoid redundancy and reduce noise, we randomly sample a subset of such candidate edges as final demonstrations.



\noindent\textbf{Instruction Tuning Objective.} As shown in Figure~\ref{fig:mlaga}, we feed the LLM with a concatenation of the enriched cross-task representations $\hat{\mathbf{e}}_{v_i}$, the raw text attribute $\mathcal{T}_{v_i}$, and the CLS token from the image encoder $\mathbf{CLS}_{\text{img},v_i}$, along with the constructed demonstration prompt $\mathcal{D}_{\text{demo}}$ (see Section~\ref{demonstration} in the  Appendix for details). The LLM is trained with an auto-regressive objective:
\begin{equation}
    \mathcal{L}_{v_i} = - \sum \log P(\mathcal{T}_{\text{ans},v_i} \mid \mathcal{T}_{v_i}, \mathbf{CLS}_{\text{img},v_i}, \hat{\mathbf{e}}_{v_i}, \mathcal{D}_{demo}),
\end{equation}
where $\mathcal{T}_{\text{ans},v_i}$ is the target output for node $v_i$ depending on the task. This unified instruction tuning setup enables the model to follow multimodal, graph-structured prompts across diverse tasks.

In summary, the MMGIT framework enables MLaGA to bridge task diversity and domain heterogeneity through a principled combination of task-aware projection, cross-task knowledge sharing, and prompt-driven instruction tuning. Together with SMA, this design explicitly addresses both Challenge 1 (modality alignment) and Challenge 2 (cross-task / domain generalization), providing an effective foundation model for multimodal graph learning.

\begin{table*}[h]
\centering
\caption{\textbf{Performance of MLaGA and leading baseline models. We report accuracy (\%) for both node classification and link prediction tasks. The methods highlighted in bold indicate the best overall performance across all models, while those \underline{underlined} represent the best results among the baseline methods. }}
\label{tab:main_results}
\begin{adjustbox}{width=\textwidth}
\begin{tabular}{@{}c|c|cccc|cccc@{}}
\toprule
\multirow{2}{*}{\textbf{Model Type}} & \multirow{2}{*}{\textbf{Model}} & \multicolumn{4}{c|}{\textbf{Node Classification Accuracy (\%)}} & \multicolumn{4}{c}{\textbf{Link Prediction Accuracy (\%)}} \\
\cline{3-6} \cline{7-10}
 & & \textbf{Movies} & \textbf{Arts} & \textbf{VideoGames} & \textbf{RedditS} & \textbf{Beauty} & \textbf{Health} & \textbf{CD} & \textbf{Art500K} \\
\midrule
\multirow{3}{*}{\makecell[c]{GNN-based}} & MLP & 45.21 & 84.28 & 90.98 & 88.09 & 63.63 & 62.63 & 78.00 & 49.10\\
 & GCN & 46.97 & 76.76 & 77.28 & 89.80 & 69.27 & 67.97 & 71.07 & 76.60\\
 & GraphSAGE & 44.07 & 85.35 & 86.76 & 90.47 & 65.53 & 62.67 & 68.37 & 72.27\\
\midrule
\multirow{2}{*}{\makecell[c]{MLLM}} & Qwen2.5-VL-7B & \underline{51.66} & 88.46 & 91.86 & 88.04 & 72.43 & 79.45 & 80.90 & 62.03\\
 & LLaVA1.5-Next-7B & 42.40 & 77.07 & 66.75 & 67.12 & 50.42 & 51.15 & 51.80 & 49.98\\
\midrule
\multirow{3}{*}{\makecell[c]{Multimodal\\Graph Model}} & MMGCN & 46.79 & 86.63 & 89.30 & 90.08 & 54.60 & 55.60 & 65.22 & 50.33 \\
 & MGAT & 40.17 & 87.25 & 85.15 & 88.31 & 53.20 & 55.40 & 67.66 & 50.56 \\
 & Unigraph2 & 45.91 & 78.81 & 81.79 & \underline{93.65} & 77.38 & 82.07& 81.00 & 64.40\\
\midrule
\multirow{6}{*}{\makecell[c]{Graph LLM}} & GraphGPT & 10.19 & 57.35 & 48.04 & 42.90& 63.50 & 62.15 & 68.95 & 39.38\\
 & GraphPrompter & 46.36 & 84.40 & \underline{92.56} & 92.85 & 76.60 & 82.95 & 79.83 & 51.45\\
 & LLaGA-ND (Single Focus) & 49.42 & 88.60 & 86.08 & 93.24 & 75.67 & 82.21 & 81.36 & 80.06\\
 & LLaGA-ND (Cross-Domain)& 50.19 & \underline{88.95} & 86.35 & 92.54 & 76.45 & 82.63 & 82.69 & 80.25\\
 & LLaGA-ND (Cross-Task) & 48.70 & 87.92 & 85.85 & 92.83 & \underline{77.92} & \underline{83.15} & \underline{84.15} & \underline{81.07}\\
\midrule
Ours & \textbf{MLaGA} & \textbf{54.99} & \textbf{90.39} & \textbf{93.44} & \textbf{96.95} & \textbf{79.80} & \textbf{83.45} & \textbf{90.20} & \textbf{84.62} \\
\bottomrule
\end{tabular}
\end{adjustbox}
\end{table*}

%% file: sections/Experiments.tex
We conduct extensive experiments to answer several key research questions. \textbf{RQ1:} How does MLaGA perform compared to state-of-the-art baselines on multimodal graph learning tasks? \textbf{RQ2:} To what extent can MLaGA serve as a generalizable foundation model across diverse graph tasks and domains? \textbf{RQ3:} Is MLaGA capable of effective knowledge transfer in typical transfer learning scenarios? 
\textbf{RQ4:} Can MLaGA scale well to new tasks? \textbf{RQ5:} How do individual components contribute to the overall performance of MLaGA?
\subsection{Experiments Setup}
\noindent\textbf{Datasets.}
We train and evaluate our model on various multimodal real-world MMG datasets across several domains: Movies, Arts, Health, VideoGames, Beauty, CD and Toys from E-commerce domain~\cite{He_2016_up}; RedditS and RedditM from Social Network domain~\cite{zhu2025mosaicmodalitiescomprehensivebenchmark, liu2025graphmllmharnessingmultimodallarge}; Art500K from Artwork domain ~\cite{mao2017deepart, mao2019visual}; Goodreads of Book-comment domain~\cite{wan_2018_item, wan_2019_fine}. The specific statistical data and description for these datasets and their splits can be found in Appendix~\ref{app:a}.


\noindent\textbf{Baselines.}
We compare the performance of MLaGA with state-of-the-art models across four different categories. The first category includes Graph Neural Networks (GNNs), such as GCN \cite{GCN}, GraphSAGE\cite{hamilton2018inductiverepresentationlearninglarge}, and MLP \cite{rosenblatt1958perceptron}. The second category consists of leading Multimodal Large Language Models (MLLMs), including LLaVA-1.5-7B \cite{liu2023visualinstructiontuning} and Qwen2.5-VL-7B ~\cite{bai2025qwen25vltechnicalreport}. The third category encompasses concurrent Graph Large Language Models, such as LLaGA \cite{chen2024llagalargelanguagegraph}, GraphPrompter \cite{www24graphprompter} and GraphGPT \cite{tang2024graphgptgraphinstructiontuning}.
The last category consists of Multimodal Graph Models, such as Unigraph2 \cite{he2025unigraph2learningunifiedembedding}, MMGCN \cite{hu2021mmgcnmultimodalfusiondeep} and MGAT \cite{tao2020mgat}.

\noindent\textbf{Implementation Details.}
For the implementation of the structure-aware multimodal encoder, we jointly pre-train the encoder across all datasets, setting the learning rate to 1e-5. We sample 1-hop neighbors for each anchor node and randomly select five neighbor nodes. For the textual and visual encoders, we employ CLIP ViT-L/14. To align the dimensionality between textual tokens and visual tokens, we use a linear projection layer to map the textual tokens into the dimensional space of the visual tokens. For the unified multimodal instruction tuning, we utilize Vicuna-7B-v1.5-16K~\cite{zheng2023judgingllmasajudgemtbenchchatbot} as MLaGA's LLM backbone. For NC demonstrations, we set k to 3. For LP demonstrations, we sample 2-hop neighborhoods $\mathcal{N}(v_i)$, $\mathcal{N}(v_j)$ of $v_i$, $v_j$ and randomly select one edge from the neighborhood as the demonstration edge. We balanced the number of samples from different tasks within each batch. For GNN-based models, we simply concatenate the textual tokens and visual tokens generated by CLIP as input. Since graph LLM baselines excel in TAGs but lack the ability to align different modalities, we use the textual tokens generated by CLIP as input. More details are in Appendix~\ref{app:b} \& Appendix~\ref{app:c} \& Appendix~\ref{app:d}.


\begin{figure}[htb]\centering
\resizebox{0.35\textwidth}{!}
{\input{figures/tex/Ablation_study}}
\caption{\textbf{The impact of MLaGA w.r.t diverse training settings.}}
\label{fig:pretraining}
\vspace{-20pt}
\end{figure}

\subsection{Main Results}
To answer \textbf{RQ1}, we pre-train MLaGA on node classification and link prediction tasks across eight datasets spanning three domains: e-commerce, social networks, and artwork. Specifically, the node classification task involves the Movies, Arts, VideoGames, and RedditS datasets, while the link prediction task is conducted on Beauty, Health, CD, and Art500K. Following prior experimental setting~\cite{chen2024llagalargelanguagegraph}, we implement three different versions of LLaGA: \textit{Single Focus} (training on one dataset and task), \textit{Cross-Domain} (training on one task across multiple datasets) and \textit{Cross-Task} (training across diverse datasets and tasks). We compare MLaGA against four categories of leading baseline models. We evaluate both MLaGA and leading baselines with in-distribution datasets, and the results are presented in Table~\ref{tab:main_results}. We observe that:

\textbf{\textit{\underline{Observation 1}}} \textit{While current leading graph LLMs can generalize across domains, they lack the capability to generalize across tasks.} Although LLaGA's performance improves with larger training datasets on single tasks, it experiences a performance drop on the node classification task under cross-task and cross-dataset training. This suggests that for current graph LLMs, it remains challenging to generalize across tasks with a single model, which verifies our motivation to explore cross-domain cross-task foundation model for multimodal graphs.

\textbf{\textit{\underline{Observation 2}}} \textit{As a single model, MLaGA significantly enhances the multimodal graph reasoning capabilities of LLMs for both node classification and link prediction tasks across multiple domains.} As shown in Table~\ref{tab:main_results}, MLaGA outperforms competing methods, including GNNs, graph LLMs, multimodal graph models and MLLM methods, across all eight datasets in both tasks. While graph LLMs such as LLaGA \cite{chen2024llagalargelanguagegraph} improve upon GNN-based approaches by leveraging LLMs to align graph structures with the LLM token space, MLaGA achieves even stronger results in every setting. Notably, MLaGA exceeds the performance of leading graph LLM baselines by an average margin of $+3.04\%$ in node classification and $+3.59\%$ in link prediction. This highlights the effectiveness of MLaGA under multimodal graph reasoning scenarios.
\subsection{How Does MLaGA Generalize Across Tasks?}
To answer \textbf{RQ2}, we investigate how different training paradigms affect MLaGA’s generalization ability. Specifically, we explore three settings: \textit{Single Focus}, \textit{Data Generalization}, and \textit{Data \& Task Generalization}. In the \textit{Single Focus} setting, MLaGA is trained on one dataset for a single task. The \textit{Data Generalization} setting expands training to multiple datasets within the same task, evaluating cross-domain generalization. Since the task type remains unchanged, a single projector is still used. The \textit{Data \& Task Generalization} setting, in contrast, introduces the most comprehensive challenge: MLaGA is jointly trained on multiple datasets and tasks. To assess the effectiveness of the Cross-Task Projector, we also test a variant that uses a single projector in this multi-task, multi-domain context. As shown in Figure~\ref{fig:pretraining}, we derive the following key insight:

~\textbf{\textit{\underline{Observation 3}}} \textit{MLaGA achieves strong generalization across both tasks and domains when equipped with the Cross-Task Projector.} While using a single projector benefits link prediction in the multi-task setting due to increased data diversity, it suffers a notable performance drop on node classification, suggesting difficulty in reconciling task-specific semantics. In contrast, MLaGA with the Multi-task Cross-attention Projector not only preserves node classification accuracy comparable to the task-specific setting but also significantly improves link prediction performance, with an average gain of $+8.2\%$ across all tasks. These findings validate the importance of collaboratively leveraging task-specific and agnostic knowledge to allow effective cross-task and cross-domain generalization, highlighting the promise of MLaGA as a unified foundation model for multimodal graph learning.

\subsection{Can MLaGA Adapt to Unseen Graphs?}
In this section, we analyze MLaGA’s generalization in both in-domain and cross-domain transfer scenarios. Transfer settings refer to the process of pretraining a model across multiple datasets on specific tasks and then evaluating its performance on an entirely unseen dataset without additional fine-tuning. This setup tests the model’s capacity to generalize learned patterns and representations to new, unseen domains. In our study, we assess MLaGA’s transfer capabilities under two distinct experimental configurations. In the initial configuration, we employ the demonstration template incorporating label signals to evaluate the MLaGA’s transfer ability under a few-shot scenario. In the subsequent configuration, we withhold label information when transferring to unseen datasets, a setup referred to as the zero-shot transfer setting. 
\subsubsection{In-Domain Transfer}
In this section, we evaluate the generalization ability of MLaGA under the in-domain transfer setting. Specifically, we assess whether the model can effectively transfer the in-distribution patterns learned during pretraining to unseen datasets within the same domain. To this end, we select two datasets—Toys from the e-commerce domain and RedditM from the social network domain—that belong to domains seen during pretraining but were not used for fine-tuning. For fair comparison, all baseline models are also pretrained under the same setting. We conduct evaluations on two representative tasks: node classification and link prediction. The results are reported in Table~\ref{tab:in_domain_nc} and Table~\ref{tab:in_domain_lp}. 

\textbf{\textit{\underline{Observation 4}}} \textit{Under both graph tasks, MLaGA demonstrates remarkable generalization and in-domain transfer ability.} In Table \ref{tab:in_domain_nc}\&\ref{tab:in_domain_lp}, MLaGA consistently outperforms baseline methods by a significant margin. Although its zero-shot variant underperforms the full MLaGA in node classification and link prediction, it still surpasses LLaGA with an average improvement of +35.85\% and +42.03\% on these tasks, respectively. This result underscores MLaGA’s robust generalization to unseen in-distribution domains and tasks. Moreover, MLaGA’s superiority over its zero-shot version further validates our strategy of incorporating highly similar neighbors as demonstrations.

\begin{table}[h!]
\caption{\textbf{Node classification results in in-domain transfer.}}
\label{tab:in_domain_nc}
\begin{tabular}{ccc}
\toprule
Model                   & Toys   & RedditM  \\ \midrule
MLP                 &   2.15     &   1.39     \\
GCN                 &   6.93    &    1.57     \\
GraphSAGE           &   5.92    &    2.43      \\
LLaGA-ND (Cross-Task)           &   \underline{34.11}    &     \underline{29.65}     \\
MLaGA (zero-shot)   &   \underline{42.75}   &     \underline{43.40}     \\
MLaGA     &   \textbf{61.29}    &     \textbf{59.15}     \\ \bottomrule
\end{tabular}
\end{table}

\begin{table}[h!]
\caption{\textbf{Link prediction results in in-domain transfer.}}
\label{tab:in_domain_lp}
\begin{tabular}{ccc}
\toprule
Model                   & Toys   & RedditM  \\ \midrule
MLP                 &    44.93    &    47.00      \\
GCN                 &    48.47    &     49.57     \\
GraphSAGE           &     50.57   &      47.63    \\
LLaGA-ND (Cross-Task)            &     \underline{53.50}   &      \underline{51.95}    \\
MLaGA (zero-shot)   &     \underline{84.42}   &      \underline{65.60}    \\
MLaGA     &   \textbf{87.37}     &     \textbf{66.10}     \\ \bottomrule
\end{tabular}
\vspace{-10pt}
\end{table}

\subsubsection{Cross-Domain Transfer}
Cross-domain transfer refers to the ability of a model to generalize knowledge learned from pre-training to a different, unseen domain. This capability is essential for building robust and scalable models that perform well in real-world scenarios with diverse and evolving data distributions. To evaluate the cross-domain transfer ability of MLaGA, we conduct experiments on the Goodreads dataset from the book-comment domain, which is entirely unseen during pretraining and poses a greater generalization challenge due to its domain shift. The results, presented in Table~\ref{tab:cross_domain}, leading to the following conclusion: 

\begin{table}[h!]
\caption{\textbf{Model performance in cross-domain transfer.}}
\label{tab:cross_domain}
\begin{tabular}{ccc}
\toprule
Model                & Node Classification & Link Prediction \\ \midrule
MLP               &         4.64            &         49.20        \\
GCN               &         14.31         &           50.13      \\
GraphSAGE         &          6.74          &           52.93      \\
LLaGA-ND (Cross-Task)         &          \underline{41.92}         &            \underline{60.45}     \\
MLaGA (zero-shot) &          \underline{42.75}           &         \underline{63.15}        \\
MLaGA   &          \textbf{53.75}           &     \textbf{65.20}            \\ \bottomrule
\end{tabular}
\vspace{-10pt}
\end{table}

\textbf{\textit{\underline{Observation 5}}} \textit{MLaGA exhibits strong cross-domain transfer ability compared to both GNN-based and graph LLM-based baselines.} In cross-domain transfer settings, MLaGA achieves average improvements of +18.04\% and +3.22\% over LLaGA in the few-shot and zero-shot settings for both tasks, respectively. This further demonstrates MLaGA's powerful cross-domain and cross-task generalization capabilities as a general-purpose foundation model. 
\subsection{How Well Does MLaGA Apply to New Tasks?}
\label{sec:5.5}
To answer \textbf{RQ4}, we evaluate MLaGA's performance by adding a new task: Multimodal Generation. The goal of this task is to summarize the key information of a Wikipedia section, by its content and image. Following the settings of MMGL~\cite{yoon2023mmgl}, we randomly sample 6,000 pages from Wikiweb2m~\cite{burns2023wikiweb2mpagelevelmultimodalwikipedia} and connect the sections within the same page to build an MMG. For MMGL and Unigraph2, we follow the official settings, by using both the text and image information with a section. For LLaGA, we implement the LLaGA-General model as a baseline. The results are shown in Table~\ref{tab:generative_task}. 

\textbf{\textit{\underline{Observation 6}}} \textit{MLaGA performs well when adapting to new graph-based tasks.} In multimodal generative tasks, MLaGA achieves improvements of +22.78\% in BLEU-4 and +19.32\% in ROUGE-L over the strong baseline UniGraph2. These results suggest that MLaGA scales well to new tasks. More analysis is in Appendix~\ref{appendix:g}.

\begin{table}[h!]
\caption{\textbf{Model performance in multimodal generation task.}}
\label{tab:generative_task}
\begin{tabular}{ccc}
\toprule
Model                & BLEU-4 & ROUGE-L \\ \midrule
MMGL(section all)              &         3.78            &         17.35        \\
Unigraph2(section all)               &         4.17         &           19.62      \\
LLaGA-ND-General         &          3.39          &           15.20      \\
MLaGA   &          \textbf{5.12}           &     \textbf{23.41}            \\ \bottomrule
\end{tabular}
\vspace{-20pt}
\end{table}

\subsection{Ablation Study}
In this section, we conduct a series of experiments to answer \textbf{RQ5}.
\subsubsection{Effect of Structure-Aware Multimodal Aligner}
Figure~\ref{fig:stage1} shows the node classification performance of GCN under different input representations. Since the Structure-Aware Multimodal Aligner leverages structural signals during training, we incorporate graph information into the image \& text input features to ensure a fair comparison. Specifically, we perform singular value decomposition (SVD) on the adjacency matrix to extract structural information from the graph. We adopt the top-128 dimensions from the eigenvectors as the structural representations. We then concatenate the resulting structure embeddings with the text and image embeddings to form the final input for our baseline model. Meanwhile, we fine-tune CLIP and employ it as a baseline for comparison. We observe that: 

~\textbf{\textit{\underline{Observation 7}}} \textit{The structure-aware multimodal aligner effectively unifies tokens from different modalities in the latent space.} Specifically, even when GCN is provided solely with the multimodal representations generated by the aligner, it surpasses variants that use image-only, text-only, or both types of features as inputs. Even after fine-tuning for text-image alignment, CLIP still underperforms compared to SMA. This finding demonstrates the effectiveness of our structure-aware multimodal aligner.

\begin{figure}[htb]
\centering
\resizebox{0.28\textwidth}{!}
{\input{figures/tex/input_features}}
\caption{\textbf{The impact of different input features on GCN.}}
\label{fig:stage1}
\vspace{-15pt}
\end{figure}

\subsubsection{Effect of Multi-task Cross-attention Projector}
We conduct an ablation study to investigate the contribution of Task-specific Projectors and Cross-task Attention. To achieve this, we pre-train our model under three different types of projectors: single projector, MoE-based projectors (with task-specific adapters only) and adaptive task-aware mapper. Specifically, for the MoE-based projector, we simply route multimodal tokens to the corresponding projector based on the task type. The findings summarized in Table~\ref{tab:ablation_ATM} indicate: 

~\textbf{\textit{\underline{Observation 8}}} \textit{The proposed Task-specific Projectors and Cross-task Attention could significantly enhance the performance of MLaGA compared to the traditional projection scheme.} Compared to single-projector and MoE-based projectors, MLaGA with Task-specific Projectors achieves superior performance. The overall results confirm that all designed components contribute positively to the performance of MLaGA.

\begin{table}[h!]
\caption{Ablation study on Multi-task Cross-attention Projector. TSP indicates Task-specific Projectors and CAL is the Cross-task Attention Layer.}
\label{tab:ablation_ATM}
\begin{tabular}{cc|cc|cc}
\toprule
\multirow{2}{*}{TSP} & \multirow{2}{*}{CAL} & \multicolumn{2}{c|}{Node Classification} & \multicolumn{2}{c}{Link Prediction} \\ \cline{3-6} 
                                          &                                       & Movies               & Arts              & Beauty           & Health           \\ \hline
\xmark                                         & \xmark                                     &           52.56           &          88.72         &        74.00          &       75.28           \\
\cmark                                         & \xmark                                     &             51.30         &           88.30        &          74.32        &           73.65       \\
\cmark                                         & \cmark                                     &             54.99         &          90.39         &         79.80         &          83.45        \\ \bottomrule
\end{tabular}
\vspace{-10pt}
\end{table}

\subsubsection{Template Analysis}
To evaluate the effectiveness of the demonstration templates, we remove all demonstrations from our templates while retaining every other multimodal element. As evidenced by the results in Table~\ref{tab:demo}, we observe that:

~\textbf{\textit{\underline{Observation 9}}} \textit{Demonstrations effectively enhance MLaGA’s performance on both node classification and link prediction.} With task-specific demonstration templates, MLaGA achieves an average improvement of $+2.5\%$ on node classification. This finding confirms our motivation to include highly similar neighbors as demonstrations in the prompt template. 

\begin{table}[ht]
\caption{The impact of demonstration templates.}
\label{tab:demo}
\begin{tabular}{c|cc|cc}
\toprule
\multirow{2}{*}{Template} & \multicolumn{2}{c|}{Node Classification} & \multicolumn{2}{c}{Link Prediction} \\ \cline{2-5} 
                          & Movies              & Arts               & Beauty           & Health           \\ \hline
w/o demonstrations        & 52.60                & 88.57              & 77.89            & 80.64            \\
w/ demonstrations         & 54.99               & 90.39              & 79.80             & 83.45            \\ \bottomrule
\end{tabular}
\vspace{-10pt}
\end{table}

%% file: figures/tex/Ablation_study.tex
\begin{tikzpicture}[
    scale=1.2,
    transform shape
]

\definecolor{TF}{HTML}{b3d7ae} 
\definecolor{IF}{HTML}{7dba7f} 
\definecolor{TIFIS}{HTML}{44935b}
\definecolor{MLaGA}{HTML}{196937}

  \begin{axis}[
      width=14cm, height=12cm,
      ymin=0, ymax=100,
      ylabel={Accuracy (\%)},
      ybar=0pt,
      bar width=12pt,
      ymajorgrids=true,
      minor y tick num=4,
      grid style={dashed,gray!30},
      symbolic x coords={
        Movies (NC),
        Arts (NC),
        Beauty (LP),
        Health (LP),
        Avg Performance
      },
      xtick=data,
      enlarge x limits=0.2,
      nodes near coords,
      every node near coord/.append style={
        font=\Large, rotate=90, anchor=west
      },
      ticklabel style={font=\large},
      label style={font=\LARGE},
      legend style={
        at={(0.5,1.05)}, anchor=south,
        legend columns=2,
        /tikz/every even column/.append style={column sep=1em},
        draw=none, font=\Large
      },
  ]

    \addplot[ybar, fill=TF!50, draw=none] coordinates {
      (Movies (NC),           49.42)
      (Arts (NC),             89.79)
      (Beauty (LP),           72.87)
      (Health (LP),           73.20)
      (Avg Performance,   71.32)
    };
    \addplot[ybar, fill=IF!50, draw=none] coordinates {
      (Movies (NC),           54.63)
      (Arts (NC),             90.51)
      (Beauty (LP),           73.55)
      (Health (LP),           73.58)
      (Avg Performance,   73.07)
    };
    \addplot[ybar, fill=TIFIS!80, draw=none] coordinates {
      (Movies (NC),           52.56)
      (Arts (NC),             88.72)
      (Beauty (LP),           74.00)
      (Health (LP),           75.28)
      (Avg Performance,   72.64)
    };
    \addplot[ybar, fill=MLaGA!80, draw=none] coordinates {
      (Movies (NC),           54.99)
      (Arts (NC),             90.39)
      (Beauty (LP),           79.80)
      (Health (LP),           83.45)
      (Avg Performance,   77.16)
    };

    \legend{
      Single Focus,
      Data Generalization,
      Data \& Task Generalization (single projector),
      Data \& Task Generalization (MLaGA)
    }
  \end{axis}
\end{tikzpicture}

%% file: figures/tex/input_features.tex
\begin{tikzpicture}[
    scale=1.5,
    transform shape,
]

\definecolor{TF}{HTML}{dae6f1}  
\definecolor{IF}{HTML}{b3cede}  
\definecolor{TIF}{HTML}{78aac8}  
\definecolor{TIFIS}{HTML}{4884af}  
\definecolor{CLIP}{HTML}{225b91}
\definecolor{OurGreen}{HTML}{153d64}

    \begin{axis}[
        ymin=10, ymax=95,
        ylabel={Accuracy (\%)},
        ybar=0,
        bar width=15,
        ymajorgrids=true,
        symbolic x coords={Movies, Arts},
        xtick=data,
        enlarge x limits=0.5,
        minor y tick num=5,
        legend columns=1,
        xtick pos=left,
        width=12cm, height=9cm,  
        legend style={
            at={(0.0,1.0)},           
            anchor=north west,        
            legend columns=1,         
            font=\huge,
            nodes={scale=0.7, transform shape},
            legend cell align=left,
        },
        nodes near coords,          
        every node near coord/.append style={font=\Large, rotate=90, anchor=west},
        ticklabel style={font=\huge},
        label style={font=\huge},
    ]

        \addplot [ybar,
            fill=TF!50,
            draw=none,
        ] coordinates {
            (Movies, 43.78) 
            (Arts, 76.93)  
        };

        \addplot [ybar,
            fill=IF!50,
            draw=none,
        ] coordinates {
            (Movies, 43.32) 
            (Arts, 74.48)  
        };

        \addplot [ybar,
            fill=TIF!50,
            draw=none,
        ] coordinates {
            (Movies, 46.97) 
            (Arts, 76.76)  
        };
        \addplot [ybar,
            fill=TIFIS!80,
            draw=none,
        ] coordinates {
            (Movies, 46.48)
            (Arts, 77.06)
        };
        
        \addplot [
            ybar,
            fill=CLIP!75,
            draw=none,
        ] coordinates {
            (Movies, 47.88) 
            (Arts, 76.52)  
        };

        \addplot [
            ybar,
            fill=OurGreen!75,
            draw=none,
        ] coordinates {
            (Movies, 48.93) 
            (Arts, 79.43)  
        };

        \legend{Text-Feature, Image-Feature, Text \& Image-Feature, Text \& Image-Feature + Structure,
        Fine-tuned CLIP,
        Ours}
    \end{axis}
\end{tikzpicture}

%% file: sections/Conclusion.tex
We introduced MLaGA, a unified foundation model that extends LLM-based reasoning to multimodal graphs (MMGs). MLaGA tackles two key challenges: fine-grained cross-modal alignment and cross-task generalization. To address these, we propose a two-stage framework comprising the Structure-Aware Multimodal Aligner (SMA) for robust, structure-aware representation learning, and the Multi-Task Multimodal Graph Instruction Tuning (MMGIT) framework for scalable and adaptive instruction tuning across tasks and domains. Extensive experiments on eight MMG benchmarks demonstrate MLaGA’s strong performance and generalization as a single model across both node classification and link prediction. Looking forward, extending MLaGA to additional modalities such as video presents a promising direction for future research.

%% file: sections/Appendix.tex
\section{Dataset Statistics}
\label{app:a}

\noindent\textbf{E-commerce Datasets.} In the e-commerce dataset, each node represents a product sold on Amazon. For each node, there is a text description and an image related to it. Edges indicate that the two products are purchased or reviewed together by the same user.

\noindent\textbf{Social Network Datasets.} In the social network dataset, each node represents a post on Reddit. For each node, there is a text comment to this post and an image related to the post. Edges indicate the two posts are commented by the same user.

\noindent\textbf{Book Datasets.}
Each node indicates a book on Goodreads. Each node contains a user comment and an image. Edges indicate that the two books are commented on by the same user.

\noindent\textbf{Artwork Dataset.}
 Following previous work \cite{jin2024instructg2isynthesizingimagesmultimodal}, we construct the Art500K dataset from the original Art500K toy dataset~\cite{mao2019visual} by connecting nodes with the same author or the same type. Each node is an artwork, where the text is the title of the artwork and the image is the artwork.

\noindent\textbf{Wikipedia Datasets.} Each node represents a section of a Wikipedia page. Edges indicate that two sections are on the same page. Each node contains a text attribute describing the section and an associated image.

\begin{table*}[htbp!]
\scriptsize
\centering
\label{table:datasets}
\caption{Dataset Statistics}
\begin{tabular}{cccccc}
\toprule
Datasets   & \#Nodes & \#Edges  & \# Classes& Split Ratio & Domain     \\ \hline
Movies     & 16,672  & 218,390 & 20 & 60/20/20    & E-commerce           \\
Toys       & 20,695  & 126,886 & 18 & 60/20/20    & E-commerce           \\
VideoGames & 13,037  & 232,503 & 6 & 60/20/20    & E-commerce              \\
Arts       & 28,195  & 197,428 & 7 & 60/20/20    & E-commerce            \\
Health       & 73,182  & 1,435,895 & 9 & 60/20/20    & E-commerce  \\ 
Beauty       & 87,465  & 1,841,368 & 9 & 60/20/20    & E-commerce  \\ 
CD       & 36,272  & 844,878 & 15 & 60/20/20    & E-commerce  \\ 
RedditS       & 15,894  & 566,160 & 20 & 60/20/20    & Social Network \\
RedditM       & 99,638  & 1,167,188 & 50 & 60/20/20    & Social Network \\
Goodreads       & 685,294  & 7,235,084 & 11 & 60/20/20    & Book-comment \\
Art500K       & 43,455  & 208,034,750 & 10 & 60/20/20    & Artwork \\
WikiWeb2M     & 600,000 & - & - & - &  Wikipedia\\
\bottomrule
\end{tabular}
\end{table*}

As for the dataset split, we maintain 60/20/20 across all datasets on the node classification task. For the link prediction task, we maintain a similar train and test set size as the node classification tasks. So, we randomly sample 5000 positive edges as the training set and 1600 positive edges as the test set for Beauty, Health, CD, and Art500K. For the generative task, we randomly sample 5000 pages as the train set, which contains 5026 sections in total.

\section{Baselines}
\label{app:b}
To conduct a comprehensive evaluation, we include three different categories of methods for assessment. Here is a brief introduction to all the baselines involved.

\paragraph{\textbf{GNN-based Models:}} 
The first category consists of GNNs for representation learning. Specifically, we evaluate:
\begin{itemize}
    \item \textbf{MLP~\cite{rosenblatt1958perceptron}}: A Multi-Layer Perceptron that serves as a non-graph-based baseline. We set it as one single linear layer.
    \item \textbf{GCN~\cite{GCN}}: A Graph Convolutional Network that captures neighborhood information using spectral convolutions. We sample neighbor nodes of the target node from two hops, selecting 15 neighbors from each node at every hop.
    \item \textbf{GraphSAGE~\cite{hamilton2018inductiverepresentationlearninglarge}}: A framework that learns node embeddings by sampling and aggregating neighboring features. The neighbor sampling policy remains the same with the GCN method.
\end{itemize}

\paragraph{\textbf{Vision-Language Models:}}  
The second category includes the state-of-the-art MLLM that integrates textual and visual representations. Specifically, we evaluate:  
\begin{itemize}  
    \item \textbf{LLaVA-1.5-7B}: A vision-language model designed for multimodal reasoning and understanding. We fine-tune LLaVA using LoRA~\cite{hu2021loralowrankadaptationlarge}, following the implementations from \url{https://github.com/haotian-liu/LLaVA}. 
    \item \textbf{Qwen2.5-VL}: A multimodal language model designed for multimodal reasoning and understanding. We fine-tune QwenVL using LoRA, following the implementations from \url{https://github.com/QwenLM/Qwen-VL}. 
\end{itemize}

\paragraph{\textbf{Multimodal Graph Models:}}
The third category includes multimodal graph models, which integrate multimodal features with GNN-based methods to enhance multimodal graph reasoning. Specifically, we evaluate:  
\begin{itemize}  
    \item \textbf{MMGCN~\cite{hu2021mmgcnmultimodalfusiondeep}}: A multimodal fused graph convolutional network designed for emotion recognition in conversations by effectively modeling multimodal dependencies and speaker-aware contextual information. We follow the implementation from \url{https://github.com/weiyinwei/MMGCN} in our experiments.  
    \item \textbf{MGAT~\cite{tao2020mgat}}: a multimodal graph attention network for personalized recommendation that adaptively captures user preferences across different modalities by disentangling interests at the modality level using gated attention on multimodal interaction graphs. Our experiments are based on the implementation from \url{https://github.com/mm-graph-benchmark/mm-graph-benchmark/tree/main/lp/MGAT}.
    \item \textbf{Unigraph2~\cite{he2025unigraph2learningunifiedembedding}}: A framework that learns a unified representation of multimodal data with MoE-base method. Our implementation is based on \url{https://github.com/yf-he/UniGraph2}
    \item
    \textbf{MMGL~\cite{yoon2023mmgl}}: A general, systematic framework that captures information from multiple multimodal neighbors and the relational structure connecting them, which excels at multimodal generative tasks. Our implementation is based on \url{https://github.com/minjiyoon/MMGL}
\end{itemize}

\section{Implementation Details}
\label{app:c}
All experiments are conducted on Linux machine with 500G RAM, and 1 NVIDIA h100 with 80GB GPU memory. For node classification demonstrations, we select top-3 neighbors. For link prediction demonstrations, we randomly select 1 demonstration edge.
The detailed pre-training hyper-parameters
are listed in Table ~\ref{para:sma} \& ~\ref{para:umg}.

\begin{table*}[h!]
\scriptsize
\caption{Pre-training hyper-parameters for our unified multimodal graph instruction tuning}
\label{para:sma}
\begin{tabular}{ccccccccc}
\Xhline{1.5pt}
LLM           & lr   & weight\_decay & optimizer & warmup\_ratio & lr\_scheduler\_type & cross\_attention\_layer & batch\_size & epochs \\ \hline

Vicuna-7B-v1.5 & 2e-5 & 0             & adamw     & 0.03          & cosine              & 1                       & 8           & 3           \\ \Xhline{1.5pt}
\end{tabular}
\end{table*}

\begin{table*}[h!]
\scriptsize
\caption{Pre-training hyper-parameters for our structure-aware multimodal aligner}
\label{para:umg}
\begin{tabular}{ccccccc}
\Xhline{1.5pt}
lr   & Self-Attention\_layers & Cross-Attention\_layers & Cross-Attention\_frequency & batch\_size & num\_epoch & num\_neighbors \\ \hline
1e-5 & 6                      & 2                       & 3                          & 16          & 1          & 5              \\ \Xhline{1.5pt}
\end{tabular}
\end{table*}


\section{Comparison with Multimodal Fused GraphLLMs}
To ensure a fair comparison between MLaGA and GraphLLMs, we extend the inputs of text-based GraphLLMs to incorporate multimodal information as node attributes. Specifically, we implement two variants: (i) an I2T variant, where we use Qwen-VL-Chat~\cite{} to convert the image associated with each node into a textual description, and concatenate it with the original textual information of the node to form its complete attribute; and (ii) a CLIP-fused variant, where we encode the textual and visual information of each node separately using CLIP, and then pool the resulting text and image representations to obtain the final node representation. For the remaining components of GraphLLMs, we keep the same settings as in the original text-based versions. The results are reported in Table~\ref{tab:clip_fuse}.
\begin{table}[h]
\scriptsize
\caption{MLaGA vs Multimodal Fused GraphLLMs. T is the text-based GraphLLM.}
\label{tab:clip_fuse}
\begin{tabular}{c|cccc}
\toprule
Method     & Movies & Arts  & CD & Art500K \\ \hline
GraphPrompter-T & 46.36  & 84.40 & 79.83      & 51.45 \\
GraphPrompter-I2T & 46.74  & 84.86 & 81.22      & 52.83 \\
GraphPrompter-CLIP & 45.54  & 83.86 & 81.63      & 53.77 \\
LLaGA-T & 48.70  & 87.92 & 84.15      & 81.07 \\
LLaGA-I2T & 50.61  & 88.83 & 85.88      & 83.23 \\
LLaGA-CLIP & 50.19  & 88.75 & 86.72      & 83.10 \\
MLaGA      & \textbf{54.99}  & \textbf{90.39} & \textbf{90.20}      & \textbf{84.62} \\ \bottomrule
\end{tabular}
\end{table}
We observe that MLaGA consistently outperforms strong GraphLLM baselines, even when these baselines are enhanced with multimodal node attributes. Although text-based GraphLLMs can achieve positive gains by extending their inputs to incorporate multimodal attributes, MLaGA still consistently outperforms these variants, further demonstrating the effectiveness of our proposed framework for multimodal graph learning.

\section{Can MLaGA transfer to previous unseen tasks?}
In this section, we investigate whether MLaGA can transfer to tasks that are unseen during pre-training. To this end, we introduce the node property prediction task, which aims to predict the degree of a node in the graph based on its attributes. We then evaluate the zero-shot transfer ability of the model pre-trained on node classification and link prediction. Specifically, since node property prediction shares similar node-level characteristics with node classification, we reuse the node-level projector as the main projector and leverage cross-task attention to integrate complementary information from other tasks. During this stage, we do not perform any task-specific parameter tuning. For demonstration templates, we follow the same demonstration selection strategy as in node classification and use the top-3 demonstrations for prompting. We randomly sample 1,000 data points from Movies and Arts as the test set. From the results in Table~\ref{tab:task}, we observe that MLaGA can generalize to new but related graph tasks. Under the zero-shot task transfer setting, MLaGA consistently outperforms the baseline model without task-specific tuning, further demonstrating its ability to generalize to previously unseen yet related tasks.
\begin{table}[]
\scriptsize
\caption{Result of zero-shot task transfer.}
\label{tab:task}
\centering
\begin{tabular}{c|cc}
\toprule
Method & Movies & Arts  \\ \hline
LLaGA  & 24.20  & 32.60 \\
MLaGA  & \textbf{42.40}  & \textbf{47.30} \\ \bottomrule
\end{tabular}
\vspace{-10pt}
\end{table}

\section{Demonstration Templates}
\label{app:d}
\label{demonstration}
\subsection{Node Classification Templates}

\noindent\textbf{E-commerce.} \textit{Given an product sold in \{subcategory\}. The text description of this product is <text>. The image description of this product is <image>. This product is also co-purchased with other products, based on which we obtain the graph-aware feature: <graph>. The task is to classify this product into xx classes:\{labels\}. Here are top-3 similar products calculated by PageRank algorithm associated with their truth class:Product 1's multimodal features are: Text feature: <demo1\_text>, Image feature :< demo1\_image>, Graph-aware feature: < demo1\_graph>; it belongs to: \{demo1\_label\}Please tell me which class the given product should belong to by considering its own multimodal features and the given demonstrations?}

\noindent\textbf{Book.} \textit{Given a book on the Goodreads Website. The user comment is: <text>. The image of this book is: <image>. The user of this book also comments on other books, based on which we obtain the graph-aware feature: <graph>. The task is to classify this book into xx classes:\{labels\}. Here are top-k similar books calculated by PageRank algorithm associated with their truth class: Book 1's multimodal features are: Text feature: <demo1\_text>, Image feature :<demo1\_image>, Graph-aware feature: <demo1\_graph>; it belongs to: \{demo1\_label\}......
Please tell me which class the given book should belong to by considering its own multimodal features and the given demonstrations?}

\noindent\textbf{Social Network.} \textit{Given a post on the Reddit Website. The user comment is:<text>. The image of this post is: <image>. The user of this post also comments on other posts, based on which we obtain the graph-aware feature: <graph>. The task is to classify this post into xx classes: \{labels\}.Here are top-k similar posts calculated by PageRank algorithm associated with their truth class:Product 1's multimodal features are: Text feature: <demo1\_text>, Image feature :<demo1\_image>, Graph-aware feature: <demo1\_graph>; it belongs to: \{demo1\_label\}......Please tell me which class the given post should belong to by considering its own multimodal features and the given demonstrations?}

\subsection{Link Prediction Templates}

\noindent\textbf{E-commerce.} \textit{Given two products sold in \{Subcategory\}. The concatenate text description of these two products is <concat\_text>. The concatenate image description of these two products is <concat\_image>. The concatenate graph-aware feature of these two products is <concat\_graph>. We need to predict whether these two products are purchased or reviewed together. Here are top-k similar products pair calculated by PageRank algorithm associated with their truth connections: Pair 1's multimodal features are: the concatenate text description of these two products is: <demo1\_concat\_text>, the concatenate image description of these two products is: <demo1\_concat\_image>, the concatenate graph-aware feature of these two products is <demo1\_concat\_graph>; Purchased or reviewed together: \{demo1\_label\}......Please tell me whether these two products should be purchased or reviewed together by considering their multimodal features and the given demonstrations.}

\noindent\textbf{Book.} \textit{Given two books on Goodreads Website. The concatenate text description of these two books is: \{concat\_node\_text\}. The concatenate image description of these two books is: <image>. The concatenate graph-aware feature of these two books is: <graph>. We need to predict whether these two books are commented by a same user. Here is top-k similar books pairs associated with their truth connections: Pair 1's multimodal features are: the concatenate text description of these two books is : <concat\_demo1\_text>, the concatenate image description of these two books is: <concat\_demo1\_ image>, the concatenate graph-aware feature of these two books is <graph>; commented by a same user: \{demo1\_label\}......Please tell me whether these two books are commented by a same user by considering their multimodal features and the given demonstrations.}

\noindent\textbf{Social Network.} \textit{Given two posts on Reddit Website. The concatenate text description of these two posts is: <concat\_text>. The concatenate image description of these two posts is: <concat\_image>. The concatenate graph-aware feature of these two posts is: <concat\_graph>. We need to predict whether these two posts are commented by a same user. Here is top-k similar posts pair associated with their truth connections: Pair 1's multimodal features are: the concatenate text description of these two posts is : <demo1\_concat\_text>, the concatenate image description of these two posts is: <demo1\_concat\_image>, the concatenate graph-aware feature of these two posts is < demo1\_concat\_graph>. Commented by a same user: \{demo1\_label\}......Please tell me whether these two products are commented by a same user by considering their multimodal features and the given demonstrations.}

\noindent\textbf{Artwork.} \textit{Given two Artworks. The concatenate text description of these two artworks is <concat\_text>. The concatenate image description of these two artworks is <concat\_image>. The concatenate graph-aware feature of these two artworks is <concat\_graph>. We need to predict whether these two artworks share the same author or the same type. Here are top-k similar artworks pair calculated by PageRank algorithm associated with their truth connections:Pair 1's multimodal features are: the concatenate text description of these two artworks is : <demo1\_concat\_text>, the concatenate image description of these two artworks is: <demo1\_concat\_image>, the concatenate graph-aware feature of these two artworks is <demo1\_concat\_graph>; Sharing same author or same type: \{demo1\_label\}. Please tell me whether these two artworks share the same author or the same type by considering their multimodal features and the given demonstrations.}
\subsection{Section Summary Templates}
\label{app:new_task}
\noindent\textbf{Wikipedia.} \textit{Given a section within a Wikipedia Page. The text description of this section is <text>. The image description of this section is <image>. The same page also contains other sections, forming a graph where the edges between sections indicate they are within the same page. The graph-aware feature of this section is <graph>. The task is to summary the central node: Here is top-1 similar section calculated by PageRank algorithm associated with their official summary:Section 1's multimodal features are: Text feature: $<demo1\_text>$, Image feature :$<demo1\_image>$, Graph-aware feature: $<demo1\_text>$; Summary of this section: $<demo1\_summary>$.Please summarize the key information of this section by considering its own multimodal features and the given demonstrations.}